\pdfoutput=1

\documentclass[11pt]{article}

\usepackage[]{acl}

\usepackage{times}
\usepackage{latexsym}

\usepackage{graphicx}
\usepackage{amsmath}
\usepackage{amssymb} 
\usepackage{multirow}
\usepackage{array}
\usepackage{booktabs} 
\usepackage{hyperref}
\usepackage{multirow}
\usepackage{subcaption}
\usepackage{caption}

\usepackage{float}

\usepackage[T1]{fontenc}

\usepackage[utf8]{inputenc}

\usepackage{microtype}

\newcommand\tf[1]{\textbf{#1}}

\title{Multi-View Document Representation Learning for Open-Domain \\
Dense Retrieval}

\author{Shunyu Zhang$^{1,}$\thanks{\quad Work done during internship at Microsoft Research Asia.}, Yaobo Liang$^{2}$, Ming Gong$^{3}$, Daxin Jiang$^{3}$, Nan Duan$^{2}$\\
  $^1$ Intelligent Computing and Machine Learning Lab, School of ASEE,  Beihang University \\
  $^2$ Microsoft Research Asia  \qquad   $^3$ Microsoft STC Asia \\
\tt $^1$zhangshunyu@buaa.edu.cn \\
\tt $^{2,3}$\{yalia, migon, djiang, nanduan\}@microsoft.com
}

\date{}

\begin{document}
\maketitle

\begin{abstract}

Dense retrieval has achieved impressive advances in first-stage retrieval from a large-scale document collection, which is built on bi-encoder architecture to produce single vector representation of query and document. However, a document can usually answer multiple potential queries from different views. So the single vector representation of a document is hard to match with multi-view queries, and faces a semantic mismatch problem. This paper proposes a multi-view document representation learning framework, aiming to produce multi-view embeddings to represent documents and enforce them to align with different queries. First, we propose a simple yet effective method of generating multiple embeddings through viewers. Second, to prevent multi-view embeddings from collapsing to the same one, we further propose a global-local loss with annealed temperature to encourage the multiple viewers to better align with different potential queries. Experiments show our method outperforms recent works and achieves state-of-the-art results.

\end{abstract}

\section{Introduction}
\label{sec:intro}

Over the past few years, with the advancements in pre-trained language models \cite{devlin-etal-2019-bert,roberta2019}, dense retrieval has become an important and effective approach in open-domain text retrieval~\cite{DPR2020, lee-etal-2019-latent,qu-etal-2021-rocketqa,ance2020}. A typical dense retriever usually adopts a bi-encoder~\cite{huang2013learning,reimers-gurevych-2019-sentence} architecture to encode input query and document into a single low-dimensional vector (usually CLS token), and computes the relevance scores between their representations. In real-world applications, the embedding vectors of all the documents are
pre

\begin{figure}[H]
\setlength{\abovecaptionskip}{0.3cm}
  \centering
  \includegraphics[width=0.45\textwidth]{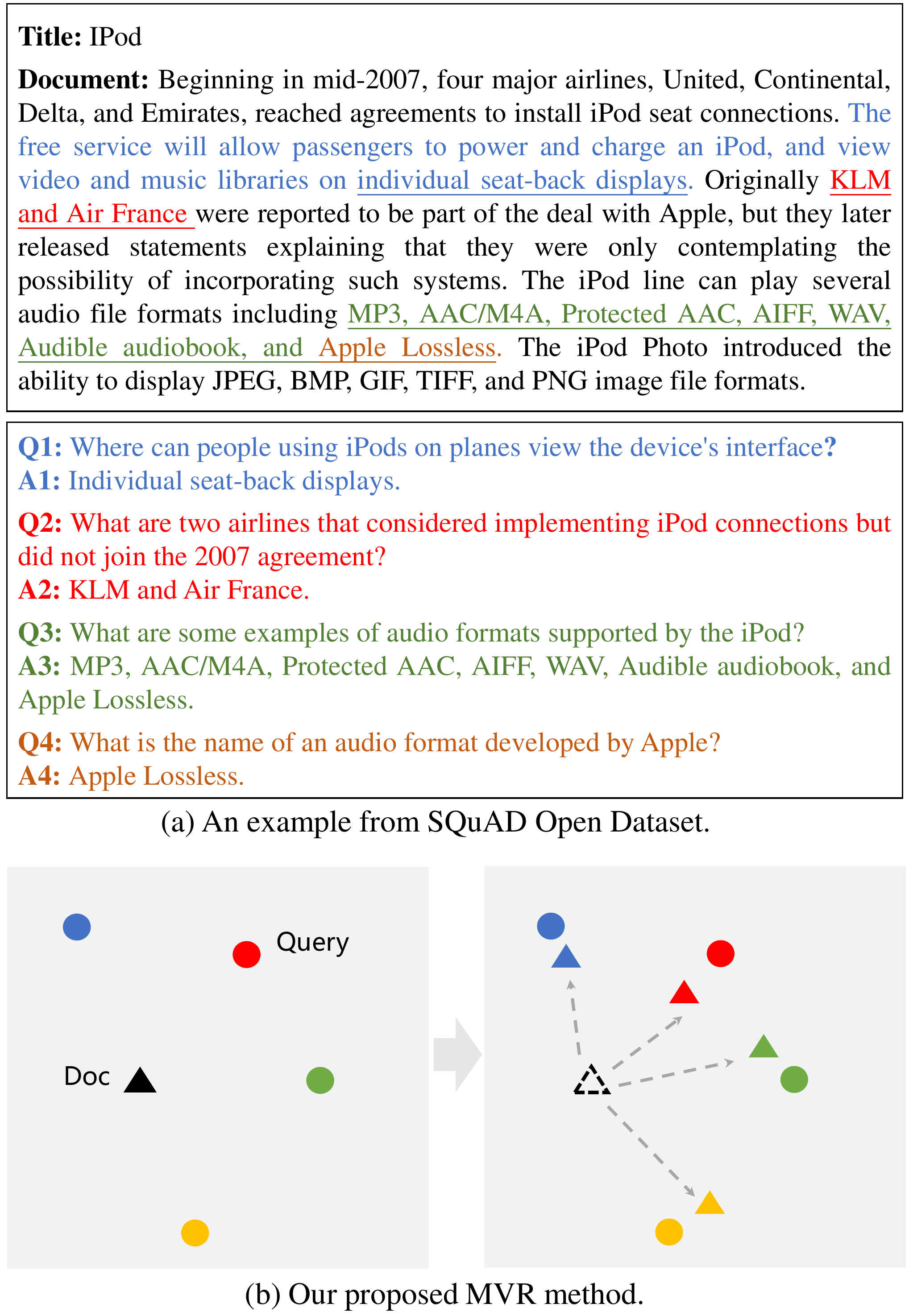}
  \caption{The illustration of our multi-view document representation learning framework. The triangles and circles mean document and query vectors separately. Usually, one document can be asked to different potential queries from multiple views. Our method comes from this observation and generates multi-view representations for documents to better align with different potential queries.}
  \label{fig:idea}
\end{figure}

\noindent -computed in advance, and the retrieval process can be efficiently boosted by the approximate nearest neighbor (ANN) technique~\cite{faiss17}.
To enhance bi-encoder's capacity, recent studies carefully design sophisticated methods to train it effectively, including constructing more challenging hard negatives~\cite{zhan2021optimizing, ance2020,qu-etal-2021-rocketqa} and continually pre-train the language models~\cite{gao2021condenser, ouguz2021dprpaq} for a better transfer.

However, being limited to the single vector representation, bi-encoder faces the upper boundary of representation capacity according to theoretical analysis in ~\citet{MEBERT2020}. 
In the real example from SQuAD dev dataset, we also find that a single vector representation can't match well to multi-view queries, as shown in Figure.\ref{fig:idea}. The document corresponds to four different questions from different views, and each of them matches to different sentences and answers. In the traditional bi-encoder, the document is represented to a single vector while it should be recalled by multiple diverse queries, which limits the capacity of the bi-encode. 

As for the multi-vector models, cross-encoder architectures perform better by computing deeply-contextualized representations of query-document pairs, but are computationally expensive and impractical for first-stage large-scale retrieval~\cite{reimers-gurevych-2019-sentence,poly2019}. Some recent studies try to borrow from cross-encoder and extend bi-encoder by employing more delicate structures, which allow the multiple vector representations and dense interaction between query and document embeddings. including late interaction~\cite{Colbert2020} and attention-based 
aggregator~\cite{poly2019,tang-etal-2021-improving-document}. However, most of them contain softmax or sum operators that can’t be decomposed into max over inner products, and so fast ANN retrieval can't be directly applied. 



Based on these observations, we propose \textit{\textbf{M}ulti-\textbf{V}iew document \textbf{R}epresentations learning} framework, \textit{\textbf{MVR}} in short. 
MVR originates from our observation that a document commonly has several semantic units, and can answer multiple potential queries which contain individual semantic content. It is just like given a specified document, different askers raise different questions from diverse views. Therefore, we propose a simple yet effective method to generate multi-view representations through \textbf{viewers}, optimized by a global-local loss with annealed temperature to improve the representation space.

Prior work has found \emph{[CLS]} token tends to aggregate the overall meaning of the whole input segment~\cite{kovaleva2019revealing,clark-etal-2019-bert}, which is inconsistent with our goal of generating multi-view embeddings. So we first modify the bi-encoder architecture, abandon \emph{[CLS]} token and add multiple \emph{[Viewer]} tokens to the document input. The representation of the viewers in the last layer is then used as the multi-view representations. 

To encourage the multiple viewers to better align with different potential queries, we propose a global-local loss equipped with an annealed temperature.
In the previous work, the contrastive loss between positive and negative samples is widely applied~\cite{DPR2020}. Apart from global contrastive loss, we formulate a local uniformity loss between multi-view document embeddings, to better keep the uniformity among multiple viewers and prevent them from collapsing into the same one. In addition, we adopt an annealed temperature which gradually sharpens the distribution of viewers, to help multiple viewers better match to different potential queries, which is also validated in our experiment.

The contributions of this paper are as follows:
\begin{itemize}
    \item We propose a simple yet effective method to generate multi-view document representations through multiple viewers.
    \item To optimize the training of multiple viewers, we propose a global-local loss with annealed temperature to make multiple viewers to better align to different semantic views.
    \item Experimental results on open-domain retrieval datasets show that our approach achieves state-of-the-art retrieval performance. Further analysis proves the effectiveness of our method.
\end{itemize}

\section{Background and Related Work}

\subsection{Retriever and Ranker Architecture}
\label{sec:architecture}
\begin{figure*}
    \centering
    \includegraphics[width=\linewidth]{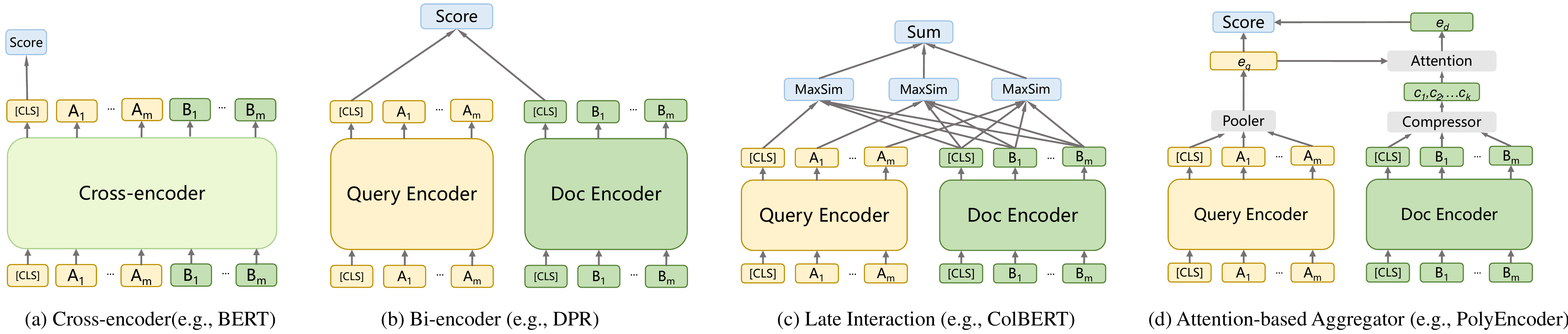}
    \caption{The comparison of different model architectures designed for retrieval/re-ranking.}
    \label{fig:comparison}
\end{figure*}

With the development of deep language model~\cite{devlin-etal-2019-bert}, fine-tuned deep pre-trained BERT achieve advanced re-ranking performance~\cite{dai2019deeper, nogueira2019passage}. The initial approach is the cross-encoder based re-ranker, as shown in Figure.\ref{fig:comparison}(a). It feeds the concatenation of query and document text to BERT and outputs the \emph{[CLS]} token's embeddings to produce a relevance score. Benefiting from deeply-contextualized representations of query–document pairs, the deep LM helps bridge both vocabulary mismatch and semantic mismatch. However, cross-encoder based rankers need computationally expensive cross-attention operations~\cite{Colbert2020, gao2021condenser}, so it is impractical for large-scale first-stage retrieval and is usually deployed in second-stage re-ranking.

As for first-stage retrieval, bi-encoder is the most adopted architecture~\cite{DPR2020} for it can be easily and efficiently employed with support from approximate nearest neighbor (ANN)~\cite{faiss17}. As illustrated in Figure.\ref{fig:comparison}(b), it feeds the query and document to the individual encoders to generate single vector representations, and the relevance score is measured by the similarity of their embeddings. Equipped with deep LM, bi-encoder based retriever has achieved promising performance~\cite{DPR2020}. And later studies have further improved its performance through different carefully designed methods, which will be introduced in Sec.\ref{sec:dense_retrieval}

Beside the typical bi-encoder, there are some variants\cite{mose2020,chen-etal-2020-dipair, mehri-eric-2021-example} proposing to employ dense interactions based on Bi-encoder. As shown in Fig.\ref{fig:comparison}(c), ColBERT~\cite{Colbert2020} adopts the late interaction paradigm, which computes token-wise dot scores between all the terms’ vectors, sequentially followed by max-pooler and sum-pooler to produce a relevance score. Later on, \citet{gao-etal-2021-coil} improve it by scoring only on overlapping token vectors with inverted lists. 
Another variant is the attention-based aggregator, as shown in Fig.\ref{fig:comparison}(d). It utilizes the attention mechanism to compress the document embeddings to interact with the query vector for a final relevance score. There are several works \cite{poly2019,MEBERT2020,tang-etal-2021-improving-document} built on this paradigm. Specifically, Poly-Encoder(learnt-k)~\cite{poly2019} sets $k$ learnable codes to attend them over the document embeddings. DRPQ~\cite{tang-etal-2021-improving-document} achieve better results by iterative K-means clustering on the document vectors to generate multiple embeddings, followed by attention-based interaction with query. However, the dense interaction methods can't be directly deployed with ANN, because both the sum-pooler and attention operator can't be decomposed into max over inner products, and the fast ANN search cannot be applied. So they usually first approximately recall a set of candidates then refine them by exhaustively re-ranking, While MVR can be directly applied in first-stage retrieval.

Another related method is ME-BERT\cite{MEBERT2020}, which adopts the first $k$ document token embeddings as the document representation. However, only adopting the first-k embeddings may lose beneficial information in the latter part of the document, while our viewer tokens can extract from the whole document. In Sec.\ref{sec:multi-view}, we also find the multiple embeddings in MEBERT will collapse to the same \emph{[CLS]}, while our global-local loss can address this problem. 


\subsection{Effective Dense Retrieval}
\label{sec:dense_retrieval}

In addition to the aforementioned work focusing on the architecture design, there exist loads of work that proposes to improve the effectiveness of dense retrieval.
Existing approaches of learning dense passage retriever can be divided into two categories: (1) pre-training for retrieval~\cite{chang2020pre, lee-etal-2019-latent, guu2020realm}  and (2) fine-tuning pre-trained language models (PLMs) on labeled data~\cite{DPR2020, ance2020, qu-etal-2021-rocketqa}. 

In the first category, \citet{lee-etal-2019-latent} and \citet{chang2020pre} propose different pre-training task and demonstrate the effectiveness of pre-training in dense retrievers. Recently, DPR-PAQ~\cite{ouguz2021dprpaq} proposes domain matched pre-training, while Condenser~\citep{gao2021condenser,gao2021cocondenser} enforces the model to produce an information-rich CLS representation with continual pre-training.

As for the second class, recent work~\cite{DPR2020, ance2020, qu-etal-2021-rocketqa, zhan2021optimizing} shows the key of fine-tuning an effective dense retriever revolves around hard negatives. DPR~\cite{DPR2020} adopts in-batch negatives and BM25 hard negatives. ANCE~\cite{ance2020} proposes to construct hard negatives dynamically during training. RocketQA~\cite{qu-etal-2021-rocketqa, ren2021rocketqav2} shows the cross-encoder can filter and mine higher-quality hard negatives. \citet{li2021more} and \citet{ren-etal-2021-pair} demonstrate that passage-centric and query-centric negatives can make the training more robust. It is worth mentioning that distilling the knowledge from cross-encoder-based re-ranker into bi-encoder-based retriever~\citep{sachan-etal-2021-end, izacard2020distilling, ren-etal-2021-pair, ren2021rocketqav2, zhang2021adversarial} can improve the bi-encoder's performance. Most of these works are built upon bi-encoder and naturally inherit its limit of a single vector representation, while our work modified the bi-encoder to produce multi-view embeddings, and is also orthogonal to these strategies.

\section{Methodology}

\subsection{Preliminary}

We start with a bi-encoder using BERT as its backbone neural network, as shown in Figure \ref{fig:comparison}(b). A typical Bi-encoder adopts dual encoder architecture which maps the query and document to single dimensional real-valued vectors separately. 

Given a query $q$ and a document collection $D=\{d_1,\dots,d_j,\dots,d_n\}$, dense retrievers leverage the same BERT encoder to get the representations of queries and documents. Then the similarity score $f(q,d)$ of query $q$ and document $d$ can be calculated with their dense representations:

\begin{equation}
\label{eq:sim}
    f(q, d) = sim(E_Q(q), E_D(d))
\end{equation}

\noindent Where $sim(\cdot)$ is the similarity function to estimate the relevance between two embeddings, e.g., cosine distance, euclidean distance, etc. And the inner-product on the \emph{[CLS]} representations is a widely adopted setting~\cite{DPR2020, ance2020}.

A conventional contrastive-learning loss is widely applied for training query and passage encoders supervised by the target task's training set. For a given query $q$, it computed negative log likelihood of a positive document $d^+$ against a set of negatives $\{d^-_1, d^-_2, .. d^-_l\}$. 

\begin{equation}
\label{eq:bi_loss}
\mathcal{L} = -\log \frac{e^{f(q, d^+)/\tau}}{ e^{f(q, d^+)/\tau} + \underset{l}{\sum} e^{f(q, d^-_l)/\tau} }
\end{equation}


Where $\tau$ is hyper-parameter of temperature-scaled factor, and an appropriate temperature can help in better optimization~\cite{sachan-etal-2021-end, li2021more}.

\subsection{Multi-Viewer Based Architecture}
\begin{figure*}
    \centering
    \includegraphics[width=0.75\linewidth]{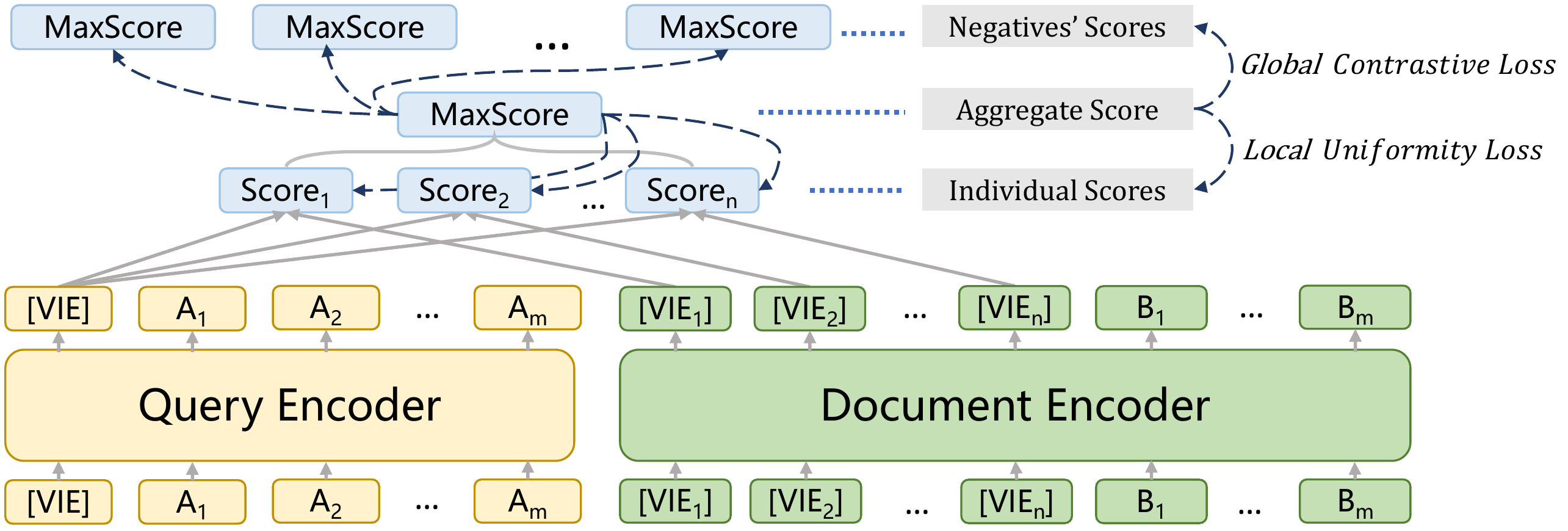}
    \caption{The general framework of multi-view representation learning with global-local loss. The gray blocks indicates the categories of scores in different layers.}
    \label{fig:modelx}
\end{figure*}

Limited to single vector representation, the typical bi-encoder faces a challenge that a document contains multiple semantics and can be asked by different potential queries from multi-view.
Though some previous studies incorporate dense interaction to allow multiple representations and somehow improve the effectiveness, they usually lead to more additional expensive computation and complicated structure.
Therefore, we propose a simple yet effective method to produce multi-view representations by multiple viewers and we will describe it in detail.

As pre-trained BERT has benefited a wide scale of the downstream tasks including sentence-level ones, some work has found \emph{[CLS]} tend to aggregate the overall meaning of the whole sentence~\cite{kovaleva2019revealing,clark-etal-2019-bert}. However, our model tends to capture more fine-grained semantic units in a document, so we introduce \textbf{multiple viewers}. Rather than use the latent representation of the \emph{[CLS]} token, we adopt newly added multiple viewer tokens \emph{[VIE]} to replace \emph{[CLS]}, which are randomly initialized. For documents input, we add different $[VIE_i]$(i=1,2,..., n) at the beginning of sentence tokens. To avoid effect on the positional encoding of the original input sentences, we set all the position ids of $[VIE_i]$ to 0, and the document sentence tokens start from 1 as the original. Then We leverage the dual encoder to get the representations of queries and documents: 

\begin{small} 
\begin{equation}
\label{eq:encode}
\begin{split}
    E(q) &= Enc_q([VIE] \circ  q \circ [SEP]) \\
    E(d) =& Enc_d([VIE_1]\cdots[VIE_n] \circ d \circ [SEP])
\end{split}
\end{equation}
\end{small}

\noindent Where $\circ$ is the concatenation operation. \emph{[VIE]} and \emph{[SEP]} are special tokens in BERT. $Enc_q$ and $Enc_d$ mean query and document encoder. We use the last layer hidden states as the query and document embeddings. 

The representations of the \emph{[VIE]} tokens are used as representations of query $q$ and document $d$, which are denoted as $E_0(q)$ and $E_i(d) (i=0,1, ..., k-1)$, respectively. As the query is much shorter than the document and usually represents one concrete meaning, we retain the typical setting to produce only one embedding for the query. 

Then the similarity score $f(q,d)$ of query $q$ and document $d$ can be calculated with their dense representations. As shown in Figure.\ref{fig:modelx}, we first compute the \emph{Individual Scores} between the single query embedding and document's multi-view embeddings, in which we adopt the inner-product. The resulted score corresponding to $[VIE_i]$ is denoted as $f_i(q,d) (i=0,1, ..., k-1)$. The we adopt a max-pooler to aggregate individual score to the \emph{Aggregate Score} $f(q,d)$ as the similarity score for the given query and document pairs:

\begin{equation}
\label{eq:maxdot}
\begin{split}
    f(q,d) &= \underset{i}{Max} \{f_i(q,d)\}   \\    
        &= \underset{i}{Max} \{sim(E_0(q), E_{i}(d))\} 
\end{split}
\end{equation}

\subsection{Global-Local Loss}

In order to encourage the multiple viewers to better align to different potential queries, we introduce a \textit{Global-Local Loss} to optimize the training of multi-view architecture. It combines the global contrastive loss and the local uniformity loss.

\begin{equation}
\label{eq:loss}
\mathcal{L} = \mathcal{L}_{global} + \lambda \mathcal{L}_{local}
\end{equation}

The global contrastive loss is inherited from the traditional bi-encoder. Given the query and a positive document $d^+$ against a set of negatives $\{d^-_1, d^-_2, .. d^-_l\}$, It is computed as follows:

\begin{equation}
\label{eq:global}
\mathcal{L}_{global} = -\log \frac{e^{f(q, d^+)/\tau}}{ e^{f(q, d^+)/\tau} + \underset{l}{\sum} e^{f(q, d^-_l)/\tau} }
\end{equation}

To improve the uniformity of multi-view embedding space, we propose applying \textit{Local Uniformity Loss} among the different viewers in Eq.\ref{eq:local}. For a specific query, one of the multi-view document representations will be matched by max-score in Eq.\ref{eq:maxdot}. The local uniformity loss enforces the selected viewer to more closely align with the query and distinguish from other viewers.

\begin{equation}
\label{eq:local}
\mathcal{L}_{local} = -\log \frac{e^{f(q, d^+)/\tau}}{\underset{k}{\sum} e^{fi(q, d^+)/\tau} }
\end{equation}

To further encourage more different viewers to be activated, we adopt an annealed temperature in Eq.\ref{eq:temperature}, to gradually tune the sharpness of viewers' softmax distribution. In the start stage of training with a high temperature, the softmax values tend to have a uniform distribution over the viewers, to make every viewer fair to be selected and get back gradient from train data. As the training process goes, the temperature decreases to make optimization more stable. 

\begin{equation}
\label{eq:temperature}
\tau = max\{0.3, exp({-\alpha t})\}
\end{equation}

Where $\alpha$ is a hyper-parameter to control the annealing speed, $t$ denotes the training epochs, and the temperature updates every epoch. To simplify the settings, we use the same annealed temperature in $\mathcal{L}_{local}$ and $\mathcal{L}_{global}$ and our experiments validate the annealed temperature works mainly in conjunction with $\mathcal{L}_{local}$ through multiple viewers. 

\begin{table*}[t]
\centering
\scalebox{0.85}{
\begin{tabular}{ l c c  c c c c c c c}
\toprule
\multirow{ 2}{*}{Method}& \multicolumn{3}{c}{\textbf{SQuAD}} & \multicolumn{3}{c}{\textbf{Natural Question}} & \multicolumn{3}{c}{\textbf{Trivia QA}}  \\
 & R@5 & R@20 & R@100 & R@5 & R@20 & R@100 & R@5 & R@20 & R@100 \\
\hline
BM25~\cite{yang2017anserini}  & -&- & - & - & 59.1 & 73.7 & - & 66.9 & 76.7 \\
DPR ~\cite{DPR2020} & -& 76.4 & 84.8 & - & 74.4 & 85.3 & - & 79.3 & 84.9 \\
ANCE~\cite{ance2020}  & -& - & - & - & 81.9 & 87.5 & - & 80.3 & 85.3\\
RocketQA~\cite{qu-etal-2021-rocketqa}  & -& - & - & 74.0 & 82.7 & 88.5 & - & - & - \\
Condenser~\cite{gao2021condenser} & - & - & - & - & 83.2 & 88.4 & - & 81.9 & 86.2\\
DPR-PAQ~\cite{ouguz2021dprpaq}  & -& - & - & 74.5 & 83.7 & 88.6 & - & - & - \\
DRPQ~\cite{tang-etal-2021-improving-document}  & -& 80.5 & 88.6 & - & 82.3 & 88.2 & - & 80.5 & 85.8\\
coCondenser~\cite{gao2021cocondenser} & - & - & - & 75.8 & 84.3 & 89.0 & 76.8 & 83.2 & 87.3\\
coCondenser(reproduced) & 73.2 & 81.8 & 88.7 & 75.4 & 84.1 & 88.8 & 76.4 & 82.7 & 86.8\\
\hline
MVR & \textbf{76.4} & \textbf{84.2} & \textbf{89.8} & \textbf{76.2} & \textbf{84.8} & \textbf{89.3} & \textbf{77.1} & \textbf{83.4} & \textbf{87.4}\\
\bottomrule
\end{tabular}
}
\caption{Retrieval performance on SQuAD dev, Natural Question test and Trivia QA test. The best performing models are marked bold and the results unavailable are left blank.}
\label{tab:sota}
\end{table*}

During inference, we build the index of all the reviewer embeddings of documents, and then our model directly retrieves from the built index leveraging approximate nearest neighbor (ANN) technique. However, both Poly-Encoder~\cite{poly2019} and DRPQ~\cite{tang-etal-2021-improving-document} adopt attention-based aggregator containing softmax or sum operator so that the fast ANN can't be directly applied. Though DRPQ proposes to approximate softmax to max operation, it still needs to first recall a set of candidates then rerank them using the complex aggregator, leading to expensive computation and complicated procedure. In contrast, MVR can be directly applied in first-stage retrieval without post-computing process as them.  Though the size of the index will grow by viewer number $k$,  the time complexity can be sublinear in index size~\cite{andoni2018approximate} due to the efficiency of ANN technique\cite{faiss17}.

\section{Experiments}

\subsection{Datasets}
\label{sec:dataset}
\noindent{\bf Natural Questions} \cite{kwiatkowski-etal-2019-natural} is a popular open-domain retrieval dataset, in which the questions are real Google search queries and answers were manually annotated from Wikipedia.

\noindent{\bf TriviaQA } \cite{joshi-etal-2017-triviaqa} contains a set of trivia questions with answers that were originally scraped from the Web.

\noindent{\bf SQuAD Open}\cite{rajpurkar-etal-2016-squad} contains the questions and answers originating from a reading comprehension dataset, and it has been used widely used for open-domain retrieval research.

We follow the same procedure in \cite{DPR2020} to preprocess and extract the passage candidate set from the English Wikipedia dump, resulting to about two million passages that are non-overlapping chunks of 100 words. Both NQ and TQA have about 60K training data after processing and SQuAd has 70k. Currently, the authors release all the datasets of NQ and TQ. For SQuAD, only the development set is available. So we conduct experiments on these three datasets, and evaluate the top5/20/100 accuracy on the SQuAD dev set and test set of NQ and TQ. We have counted how many queries correspond to one same document and the average number of queries of SQuAD, Natural Questions and Trivia QA are 2.7, 1.5 and 1.2, which indicates the multi-view problem is common in open-domain retrieval.

\subsection{Implementation Details}

We train MVR model following the hyper-parameter setting of DPR~\cite{DPR2020}. All models are trained for 40 epochs on 8 Tesla V100 32GB GPUs. We tune different viewer numbers on the SQuAD dataset and find the best is 8, then we adopt it on all the datasets. To make a fair comparison, we follow coCondenser~\cite{gao2021cocondenser} to adopt mined hard negatives and warm-up pre-training strategies, which are also adopted in recent works~\cite{ouguz2021dprpaq, gao2021condenser} and show promotion. Note that we only adopt these strategies when compared to them, while in the ablation studies our models are built only with the raw DPR model. During inference, we apply the passage encoder to encode all the passages and index them using the Faiss IndexFlatIP index~\cite{faiss17}.


\subsection{Retrieval Performance}

We compare our MVR model with previous state-of-the-art methods. Among these methods, DRPQ~\cite{tang-etal-2021-improving-document} achieved the best results in multiple embeddings methods, which is the main compared baseline for our model. In addition, we also compare to the recent strong dense retriever, including ANCE~\cite{ance2020}, RocekteQA~\cite{qu-etal-2021-rocketqa}, Condenser~\cite{gao2021condenser}, DPR-PAQ~\cite{ouguz2021dprpaq} and coCondenser~\cite{gao2021cocondenser}. 
For coCondenser, we reproduced its results and find it a little lower than his reported one, maybe due to its private repo and tricks. Overall, these methods mainly focus on mining hard negative samples, knowledge distillation or pre-training strategies to improve dense retrieval. And our MVR framework is orthogonal to them and can be combined with them for better promotion.

As shown in Table \ref{tab:sota}, we can see that our proposed MVR performs better than other models. Compared to DRPQ which performs best in the previous multi-vector models, MVR can outperform it by a large margin, further confirming the superiority of our multi-view representation. It's worth noting that our model improves more on the SQuAD dataset, maybe due to the dataset containing more documents that correspond to multiple queries as we state in Sec.\ref{sec:dataset}. It indicates that MVR can address the multi-view problem better than other models.


\begin{table}[t]
\small
    \centering
    \resizebox{0.42\textwidth}{!}{
    \begin{tabular}{l|rrc}
    \toprule
    Models   & \multicolumn{1}{c}{R@5} & \multicolumn{1}{c}{R@20} & \multicolumn{1}{c}{R@100} \\ 
    \hline
    DPR(k=1) & 66.2                   & 76.8                    & 85.2                     \\
    ME-BERT(k=4) & 66.8                   & 77.6                    & 85.5                     \\
    ME-BERT(k=8) & 67.3                   & 77.9                   & 86.1                   \\
    \hline
    MVR(k=4)      & 68.5                   & 78.5                    & 85.8                     \\
    MVR(k=6)      & 72.3                   & 80.3                    & 86.4                     \\
    MVR(k=8)      & \textbf{75.5}                   & \textbf{83.2}                    & \textbf{87.9}                     \\
    MVR(k=12)     & 74.8                   & 82.9                    & 87.4                     \\ 
    \bottomrule
    \end{tabular}
    }
    \caption{Performance of different viewers' number in MVR and compared models.}
    \label{tab:viewer}
\end{table}

\begin{table}[t]
    \centering
    \resizebox{0.45\textwidth}{!}{
    \begin{tabular}{cl|rrc}
    \toprule
     & Models   & \multicolumn{1}{c}{R@5} & \multicolumn{1}{c}{R@20} & \multicolumn{1}{c}{R@100} \\  \hline
    (0) & MVR ($\alpha=0.1$)  & \textbf{75.5}   & \textbf{83.2}                    & \textbf{87.9} \\
    (1) & w/o LC loss & 73.7                   & 82.1                   & 86.5\\ 
    (2) & w/o Annealed $\tau$(Fixed=1) & 74.3                   & 81.9                    & 86.8 \\
    (3) & w/o LC loss + Annealed $\tau$ & 72.8                   & 81.0     & 85.8 \\
    (4) & w/o Multiple Viewers & 66.7                  & 77.1     & 85.7 \\
    \hline
    (5) & Fixed $\tau=10$   & 70.2                   & 79.3                    & 85.4\\
    (6) & Fixed $\tau=0.3$  & 74.6                   & 82.4                    & 87.3\\
    (7) & Fixed $\tau=0.1$  & 72.3                   & 81.2                   & 85.9\\ 
    (8) & Annealed $\tau$($\alpha=0.3$)  & 74.7                   & 82.0                   & 87.4\\ 
   (9) & Annealed $\tau$($\alpha=0.03$)  & 73.9                   & 81.8                   & 86.5\\ 
    \bottomrule
    \end{tabular}
    }
    \caption{Ablation study on Global-local Loss on SQuAD dev set.}
    \label{tab:loss}
\end{table}

\subsection{Ablation Study}

\hspace{1em}{\bf Impact of Viewers' Number:} We conduct ablation studies on the development set of SQuAD open. For fair comparison, we build all the models mentioned in the following based on DPR toolkit, including MEBERT and MVR. The results are shown in Table \ref{tab:viewer}, and the first block shows the results of DPR and MEBERT which adopt the first $k$ token embeddings. Compared to DPR and MEBERT, our model shows strong performance, which indicates the multi-view representation is effective and useful.
Then, we analyze how the different numbers of viewers ($k=4,6,8,12$) affect performance in MVR. We find that the model achieves the best performance when $k=8$. When k increase to $k=12$ or larger, it leads little decrease in the performance, maybe due to there being not so many individual views in a document.

\vspace*{0.5\baselineskip} 
{\bf Analysis on Global-local Loss:} In this part, we conduct more detailed ablation study and analysis of our proposed Global-local Loss. As shown in Table \ref{tab:loss}, we gradually reduce the strategies adopted in our model. We find not having either local uniformity loss (LC loss in table) or annealed temperature damages performance, and it decreases more without both of them. We also provide more experimental results to show the effectiveness of the annealed temperature. We first tune the fixed temperature, find it between 0.3 to 1 is beneficial. We adopt the annealed temperature annealed from 1 to 0.3 gradually as in Eq.\ref{eq:temperature}, finding a suitable speed($\alpha = 0.1$) can better help with optimization during training. Note that the model w/o Multiple Viewers can be seen as DPR with annealed $\tau$, just little higher than raw DPR in Table \ref{tab:viewer}, while annealed  $\tau$ improves more when using multi-viewer. It indicates our annealed strategy plays a more important role in multi-view learning.

\begin{table}[t]
\small
    \centering
    \resizebox{0.38\textwidth}{!}{
    \begin{tabular}{l|c|c}
    \toprule
     Method  & Doc Encoding & Retrieval \\
      \hline
      DPR & 2.5ms & 10ms \\
      ColBERT & 2.5ms & 320ms \\
      MEBERT & 2.5ms  & 25ms \\
      DRPQ & 5.3ms   & 45ms     \\
      MVR & 2.5ms &  25ms \\
      \bottomrule
    \end{tabular}}
    \caption{Time cost of online and offline computing in SQuAD retrieval task.}
    \label{tab:efficiency}
    \vspace{-1em}
\end{table}

\vspace*{0.5\baselineskip} 
{\bf Efficiency Analysis: } We test the efficiency of our model on 4 Nvidia Tesla V100 GPU for the SQuAD dev set, as shown in Table \ref{tab:efficiency}. We record the encoding time per document and retrieval time per query, and don't include the query encoding time for it is equal for all the models. To compare our approach with other different models, we also record the retrieval time of other related models. We can see that our model spends the same encoding time as DPR, while DRPQ needs additional time to run K-means clustering. With the support of Faiss, the retrieval time MVR cost is near to DPR and less than ColBERT~\cite{Colbert2020} and DRPQ~\cite{tang-etal-2021-improving-document} which need additional post-computing as we state in Sec.\ref{sec:architecture}.

\section{Further Analysis}

\subsection{Comparison to Sentence-level Retrieval}

To analyze the difference between MVR and sentence-level retrieval which is another way to produce multiple embeddings, we design several models as shown in Table \ref{tab:split}. Sentence-level means that we split all the passages into individual sentences with NLTK toolkit\footnote{\url{www.nltk.org}}, resulting to an average of 5.5 sentences per passage. The new positives are the sentences containing answers in the original positives, and new negatives are all the split sentences of original negatives. K-equal-splits means the DPR's original 100-words-long passages are split into $k$ equal long sequences and training positives and negatives as Sentence-level's methods. Compared to MVR, Sentence-level drops a lot even lower than DPR maybe for they lose contextual information in passages. It also indicates that the multi-view embeddings of MVR do not just correspond to sentence embeddings, but capture semantic meanings from different views. For even a single sentence may contain diverse information that can answer different potential queries (as in Fig.\ref{fig:idea}). The k-equal-split methods perform worse much for it further lose the sentence structure and may contain more noise.


\begin{table}[t]
\small
    \centering
    \resizebox{0.45\textwidth}{!}{
    \begin{tabular}{l|ccc}
    \toprule
    Models   & \multicolumn{1}{c}{R@5} & \multicolumn{1}{c}{R@20} & \multicolumn{1}{c}{R@100} \\ 
    \hline
    DPR & 66.2                   & 76.8                    & 85.2                 \\
    MVR      & \textbf{75.5}                   & \textbf{83.2}     & \textbf{87.9}\\
    Sentence-level     & 62.1                   & 73.2                    & 81.9       \\ 
    4-equal-splits     & 57.2                   & 69.3                    & 78.5       \\ 
    8-equal-splits     & 44.2                   & 57.9                    & 69.4       \\ 
    \bottomrule
    \end{tabular}
    }
    \caption{Performance of different sentence-level retrieval models on SQuAD dev.}
    \label{tab:split}
\end{table}

\subsection{Analysis of Multi-View Embeddings}
\label{sec:multi-view}

To further show the effectiveness of our proposed MVR framework, we evaluate the distribution of multi-view document representations. We conduct evaluations on the randomly sampled subset of SQuAD development set, which contains 1.5k query-doc pairs and each document has an average of 4.8 corresponding questions. We adopt two metrics \textit{Local Variation} and \textit{Perplexity}~\cite{brown-etal-1992-estimate}(denoted as \textit{LV} and \textit{PPL}) to illustrate the effect of different methods. 
We first compute the normalized scores between the document's multi-view embeddings and query embedding as in Eq.\ref{eq:maxdot}, and record the scores $f_i(q,d)$ of all the viewers. Then Local Variation of a query-doc pair can be computed as follows, and then we average it on all the pairs.

\begin{equation}
\small
\label{eq:outdistance}
LV = {Max}(f_i(q,d)) - \frac{\underset{k}{\sum} f_i(q,d) - {Max}(f_i(q,d))}{k-1} 
\end{equation}

The Local Variation measures the distance of the max scores to the average of the others, which can curve the uniformity of different viewers. The higher it is, the more diversely the multi-view embeddings are distributed.

Then we collect the index of the viewer having the max score, and group the indexes of different queries corresponding to the same documents. Next, we can get the distributions of different indexes denoted as $p_i$. The Perplexity can be computed as follows:
\begin{equation}
\label{eq:ppl}
PPL = exp(-{\underset{m}{\sum} p_i \log p_i })
\end{equation}

If different viewers are matched to totally different queries, the $p_i$ tends to be a uniform distribution and PPL goes up. 
The comparison results are shown in Table \ref{tab:distribution}. When evaluating MEBERT, we find its multiple embeddings collapse into the same [CLS] embeddings rather than using the different token embeddings. So its PPL is near to one and Local Variation is too high. For MVR model, we find that without local uniformity loss (LC loss in short), the Local Variation drops rapidly, indicating our proposed LC loss can improve the uniformity of different viewers. In addition, MVR w/o annealed $\tau$ will damage the PPL, which also confirms it does help activate more viewers and align them better with different queries.

\begin{table}[t]
\small
    \centering
    \resizebox{0.4\textwidth}{!}{
    \begin{tabular}{l|cc}
    \toprule
    Models   & \multicolumn{1}{c}{PPL} & \multicolumn{1}{c}{LV}  \\ 
    \hline
    ME-BERT & 1.02                   & 0.159                                     \\
    MVR     & 3.19                   & 0.126                                 \\
    MVR w/o LC loss      & 3.23              &       0.052      \\           
    MVR w/o Annealed $\tau$      & 2.95                   & 0.118  \\
    \bottomrule
    \end{tabular}
    }
    \caption{Analysis of multi-view embeddings produced by different methods.}
    \label{tab:distribution}
    \vspace{-1em}
\end{table}

\subsection{Qualitative Analysis}
\label{sec:qualitative}
\begin{table*}[t]
\scriptsize
\begin{tabular}{p{3cm}|p{6cm}|p{6cm}}
\toprule
\tf{Question} & \tf{Passage received by DPR} & \tf{Passage retrieved by MVR} \\
\midrule
\multirow{2}{3cm}{What continent ranged over the majority of the southern hemisphere of earth in the Cambrian?} & Title: Mesozoic  & Title: Geological history of Earth\\
& \ldots and the Khingan Mountains in Manchuria. This orogeny was related to the opening of the Arctic Ocean and subduction of the North China and Siberian cratons under the Pacific Ocean. In contrast, the era featured the dramatic rifting of the supercontinent Pangaea, which gradually split into a northern continent, Laurasia, and a southern continent, Gondwana. This created the passive continental margin that characterizes most of the Atlantic coastline (such as along the U.S. East Coast) today. By the end of the era, the continents \ldots 

& \ldots Laurentia, Baltica and Siberia remained independent continents following the break-up of the supercontinent of \textbf{Pannotia}. Gondwana started to drift toward the South Pole. Panthalassa covered most of the southern hemisphere, and minor oceans included the Proto-Tethys Ocean, Iapetus Ocean and Khanty Ocean. The Ordovician period started at a major extinction event called the Cambrian
-Ordovician extinction event some time about 485.4 ± 1.9 Ma. During the Ordovician the southern continents were collected into a single continent called Gondwana. Gondwana started the period in \ldots \\

\midrule
\multirow{2}{3cm}{How long ago did the Ordovician period begin?} & Title: Ordovician & Title: Geological history of Earth \\

& \ldots is a geologic period and system, the second of six periods of the Paleozoic Era. The Ordovician spans 41.2 million years from the end of the Cambrian Period million years ago (Mya) to the start of the Silurian Period Mya. The Ordovician, named after the Celtic tribe of the Ordovices, was defined by Charles Lapworth in 1879 to resolve a dispute between followers of Adam Sedgwick and Roderick Murchison, who were placing the same rock beds in northern Wales into the Cambrian and Silurian systems, respectively. \ldots &

\ldots Laurentia, Baltica and Siberia remained independent continents following the break-up of the supercontinent of Pannotia.  Gondwana started to drift toward the South Pole. Panthalassa covered most of the southern hemisphere, and minor oceans included the Proto-Tethys Ocean, Iapetus Ocean and Khanty Ocean. The Ordovician period started at a major extinction event called the Cambrian-Ordovician extinction event some time about \textbf{485.4 ± 1.9 Ma}. During the Ordovician the southern continents were collected into a single continent called Gondwana. Gondwana started the period in \ldots\\

\bottomrule
\end{tabular}
\caption{Examples of passages returned from DPR and MVR. Correct answers are written in \tf{bold}.}
\label{tab:app_examples}
\end{table*}

As shown in Table \ref{tab:app_examples}, there are two examples retrieved by DPR and MVR for qualitative analysis. The top scoring passages retrieved by DPR can't give a clear answer for the queries, though they seem to have a similar topic to the queries. In contrast, our MVR is able to return the correct answers when the passages contain rich information and diverse semantics. Take the second sample as an example, the passage retrieved by DPR is around \emph{Ordovician} in the question but there are no more details answering the question. In comparison, MVR mines more fine-grained information in the passage and return the correct answer \emph{485.4 ± 1.9 Ma} (Ma means million years ago).
It indicates that DPR can only capture the rough meaning of a passage from a general view, while our MVR is able to dive into the passage and capture more fine-grained semantic information from multiple views.

\section{Conclusions}

In this paper, we propose a novel Multi-View Representation Learning framework. Specifically, we present a simple yet effective method to generate multi-view document representations through multiple viewers. To optimize the training of multiple viewers, we propose a global-local loss with annealed temperature to enable multiple viewers to better align with different semantic views. We conduct experiments on three open-domain retrieval datasets, and achieve state-of-the-art retrieval performance. Our further analysis proves the effectiveness of different components in our method.  

\section{Acknowledgements}

We thank Yuan Chai, Junhe Zhao, Yimin Fan, Junjie Huang and Hang Zhang for their discussions and suggestions on writing this paper.

\bibliography{anthology,custom}
\bibliographystyle{acl_natbib}

\appendix

\section{Scale Factor of Global-Local Loss}
\label{app_sec:scale_factor}
We have tuned the scale factor $\lambda$ of the Global-local loss in Eq.\ref{eq:loss}. The performances on SQuAD dev set are shown in Table \ref{tab:lambda}. We find that a suitable scaling factor ($\lambda$=0.01) can improve more than others. Analysing other results, we infer that a large factor of local uniformity loss may lead to much impact on optimization of global loss, while a smaller one will degenerate into the form without local uniformity loss.

\begin{table}[h]
\small
    \centering
    \resizebox{0.35\textwidth}{!}{
    \begin{tabular}{c|ccc}
    \toprule
     $\lambda$  & \multicolumn{1}{c}{R@5} & \multicolumn{1}{c}{R@20} & \multicolumn{1}{c}{R@100} \\ 
    \hline
    0.5      & 72.4       & 80.4      & 85.9       \\
    0.05     & 74.7        & 82.5      & 87.3      \\ 
    0.01     & 75.5        & 83.2      & 87.9      \\ 
    0.001    & 72.9        & 82.2      & 85.7      \\ 
    \bottomrule
    \end{tabular} }
    \caption{Performance on SQuAD dev set under different setting of scale factor.}
    \label{tab:lambda}
\end{table}

\end{document}